\documentclass[twoside]{article}

\usepackage[accepted]{aistats2026}
\usepackage{natbib}
 
\usepackage{amsthm}
\newtheorem{lemma}{Lemma}
\usepackage{needspace}

\usepackage{float}  

\usepackage[utf8]{inputenc} 
\usepackage[T1]{fontenc}    
\usepackage{hyperref}       
\usepackage{url}            
\usepackage{booktabs}       
\usepackage{amsfonts}       
\usepackage{amsmath}        
\usepackage{nicefrac}       
\usepackage{microtype}      
\usepackage{xcolor}         

\usepackage{algorithmic}

\usepackage{amsthm}

\usepackage{dsfont}       
\usepackage{subcaption}   
\usepackage[capitalize]{cleveref}
\usepackage[hybrid]{markdown}
\usepackage{zref}
\usepackage{etoolbox}
\usepackage{algorithm}
\usepackage{tikz}
\usepackage{xcolor}
\usepackage{makecell}
\usepackage{pifont}  
\usepackage{mathtools} 

\providecommand{\greentick}{\textcolor[HTML]{009B55}{\ding{51}}}
\providecommand{\redtick}{\textcolor[HTML]{FF0000}{\ding{51}}}
\providecommand{\greencross}{\textcolor[HTML]{009B55}{\ding{55}}}
\providecommand{\redcross}{\textcolor[HTML]{FF0000}{\ding{55}}}

\providecommand*{\B}[1]{\ifmmode\bm{#1}\else\textbf{#1}\fi}

\providecommand*{\Z}[1]{\mathds{#1}}

\providecommand*{\inputspace}{\mathcal{X}}
\providecommand*{\outputspace}{\mathcal{Y}}

\providecommand*{\eqn}[1]{\begin{align}#1\end{align}}

\newtoggle{include-notes}
\toggletrue{include-notes}

\definecolor{darkred}{rgb}{0.85, 0.0, 0.0}

\usepackage{amsthm}
\newtheorem{thm}{Theorem}
\newtheorem*{thm*}{Theorem}

\newtheorem{defn}[thm]{Definition}

\crefname{defn}{Definition}{Definitions}
\crefname{equation}{}{}

\crefname{defn}{definition}{definitions}
\Crefname{defn}{Definition}{Definitions}

\newcommand{\Backprop}{Backprop}

\newcommand{\VIJP}{\textit{vijp}}

\begin{document}

%

%

\twocolumn[

\aistatstitle{Moonwalk: Inverse-Forward Differentiation}

\aistatsauthor{ Dmitrii Krylov  \And Armin Karamzade  \And  Roy Fox  }

\aistatsaddress{ University of California, Irvine \And  University of California, Irvine \And University of California, Irvine } ]

\begin{abstract}
Backpropagation’s main limitation is its need to store intermediate activations (residuals) during the forward pass, which restricts the depth of trainable networks. This raises a fundamental question: can we avoid storing these activations? We address this by revisiting the structure of gradient computation. Backpropagation computes gradients through a sequence of vector–Jacobian products, an operation that is generally irreversible. The lost information lies in the cokernel of each layer’s Jacobian. We define submersive networks—networks whose layer Jacobians have trivial cokernels—in which gradients can be reconstructed exactly in a forward sweep without storing activations. For non-submersive layers, we introduce fragmental gradient checkpointing, which records only the minimal subset of residuals necessary to restore the cotangents erased by the Jacobian. Central to our approach is a novel operator, the vector–inverse-Jacobian product (vijp), which inverts gradient flow outside the cokernel. Our mixed-mode algorithm first computes input gradients with a memory-efficient backward pass, then reconstructs parameter gradients in a forward sweep that does not need to store activations. We implement this method, called Moonwalk, and show that it matches backpropagation’s runtime while training networks more than twice as deep under the same memory budget.
  


\end{abstract}



\section{Introduction}
\label{sec:intro}
Training deep neural networks via reverse‐mode automatic differentiation, commonly known as backpropagation (Backprop), requires storing \emph{residuals} during the forward pass—intermediate values sufficient to evaluate the vector–Jacobian products (\textit{vjp}) used in the backward pass.  This incurs a memory cost that grows with the network depth and often limits model capacity.  Existing techniques such as reversible layers~\citep{revnet} and activation checkpointing~\citep{martens2012training} modify \Backprop\ to reduce its memory footprint, but either impose significant architectural constraints or increase computation time. In contrast, forward‐mode differentiation~\citep{OldForward} does not impose any constraints on model architecture, and only takes a single forward pass, but suffers from prohibitively high computational cost.


In this paper, we show that for a broad class of \emph{submersive layers}, i.e. those whose Jacobians are surjective, 
true gradients can be computed in forward mode with significantly lower memory.
%
%
Our core insight is that once the loss gradient with respect to the network input (the input \emph{cotangent}) is known, all parameter gradients can be recovered through a sequence of right-inverse Jacobian products in a single forward sweep. 
%
%
%
This yields the \textbf{Moonwalk} method, which in some architectures can match Backprop’s runtime while using less memory. Moonwalk has two variants, \textbf{mixed-mode} and \textbf{pure-forward}, that pre-compute the input cotangent in reverse-mode and, respectively, forward-mode. 



Traditional backpropagation computes gradients in two distinct phases. In the first phase, it performs a forward pass to compute and store the residuals of all layers. These stored residuals are then used in the second phase to compute cotangents (loss gradients) in reverse topological order, from the loss variable back to the first layer, using the gradient chain rule. For large networks, the memory footprint scales with the number of residuals \citep{reducing_activations}, which can result in significant memory overhead, limiting the ability to scale neural networks.

Moonwalk first computes the input cotangent, either in reverse-mode (in mixed-mode Moonwalk) or in forward-mode (in pure-forward Moonwalk).
Because this phase only computes the input cotangent and not parameter gradients, it can, in some architectures, incur significantly less memory overhead than full automatic differentiation.
Moonwalk then performs a forward sweep using our custom vector-inverse-Jacobian product (\textit{vijp}) operator to recover parameter gradients. It offers the same computational complexity as Backprop in architectures where \textit{vijp} is as efficient as \textit{vjp}, but with a reduced memory footprint since parameter gradients are computed without storing full residuals.

In summary, this work contributes:
\vspace{-1.75ex}
\begin{itemize}
  \item A novel algebraic identity (Eq.~\ref{eq:ifp-main}) enabling true forward-mode gradient computation in submersive networks, given a pre-computed input cotangent.
  \item Two variants of the Moonwalk method, pure-forward and mixed-mode, both offering reduced memory footprints with a computational trade-off.
  \item An efficient implementation of the required \textit{vijp} operators, including derived conditions for submersive convolutions and fully parallelizable forms.
  \item Fragmental gradient checkpointing, a memory-efficient scheme for forward-mode reconstruction of complete cotangents from partially stored ones.
  
  \item Empirical results on submersive convolutional layers using efficient \VIJP\ operators, demonstrating up to $\times$2 memory reduction with comparable computation time, enabling the training of networks more than $\times$2 deeper or with larger batch sizes than backpropagation.

\end{itemize}




\section{Related Work}
\noindent\textbf{Reducing automatic differentiation memory.} Prior work has explored various strategies to reduce the memory footprint of training deep networks. A widely used approach is \emph{activation checkpointing}~\citep{martens2012training, chen2016training, gruslys2016memory, kumar, checkmate,rockmate,hiremate,mario, lynx,autohete,sharding,rear}, which reduces memory usage in a network with $L$ layers by a factor of $O(\sqrt{L})$. This is achieved by storing only $O(\sqrt{L})$ activations (layer outputs) during the forward pass and running a second forward pass inside the backward loop to rematerialize intermediate values. While effective, this technique increases compute time and still requires full residuals in the inner-loop forward pass.

\noindent\textbf{Invertible architectures.} Another line of work leverages invertible layers that allow activations to be recomputed exactly during the backward pass~\citep{revnet, mackay2018reversible, mangalam2022reversible, reversecol, reversellm}. Reversible backpropagation that avoids storing activations altogether has been applied to memory-efficient training~\citep{revnet}, improved representations~\citep{jacobsen2018revnet}, and generative models~\citep{kingma2018glow, dinh2014nice, rezende2015variational}. For example, \citet{bulo2018place} replaced ReLU and batch normalization layers with invertible alternatives, reducing memory usage by up to 50\%. Synthetic gradients have also been explored to decouple gradient computation~\citep{jaderberg2017decoupled}. However, these methods are restricted to architectures where layers are exactly invertible. In contrast, Moonwalk applies to the larger class of \emph{submersive networks}, whose layer Jacobians are everywhere surjective but not necessarily invertible.

\noindent\textbf{Forward-mode differentiation.} Forward-mode automatic differentiation has been proposed as an alternative to reverse-mode, particularly in recurrent networks (as in RTRL~\citep{OldForward}). More recently, projection-based variants have emerged, where directional derivatives are used to approximate true gradients~\citep{silver2021learning, baydin2022gradients}. While these approaches reduce memory costs, they introduce gradient noise due to stochastic tangent vectors or imprecise surrogate networks~\citep{ren2022scaling, forwardmatchback}, leading to subpar optimization performance. In contrast, Moonwalk avoids both full Jacobian materialization and projection noise by computing exact gradients using an efficient forward-mode operator, the vector–inverse-Jacobian product (\textit{vijp}).

\noindent\textbf{Vector-Inverse-Jacobian product (vijp).} 
While the operator itself is not typically named explicitly, closely related inverse-Jacobian computations appear in several adjacent lines of work. In particular, inverse accumulation–mode automatic differentiation formulates gradient propagation via inverse-Jacobian applications rather than standard vector–Jacobian products, providing a principled alternative to reverse-mode AD~\citep{inverseaccum}.
Similarly, implicit differentiation methods such as Deep Equilibrium (DEQ) ~\citep{DEQ} models rely on solving linear systems involving Jacobians, where inverse-Jacobian products are central to computing gradients without unrolling the forward computation. In contrast to these approaches, which typically rely on iterative or approximate solvers, our \textit{vijp} operator enables exact and efficient forward recovery of cotangents in submersive networks, forming the basis of the Moonwalk algorithm.
\section{Background}


\begin{figure*}[t]
    \centering
    \begin{tabular}{cc}
        \begin{subfigure}[b]{0.48\textwidth}
            \centering
            \includegraphics[width=\textwidth]{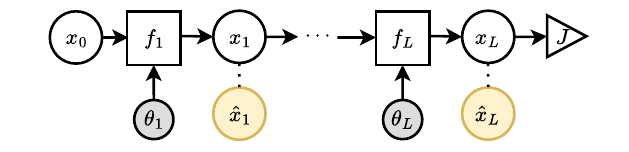}
            \caption{Moonwalk: Forward pass (Phase I)}
            \label{fig:h0_fp}
        \end{subfigure}
        &
        \begin{subfigure}[b]{0.48\textwidth}
            \centering
            \includegraphics[width=\textwidth]{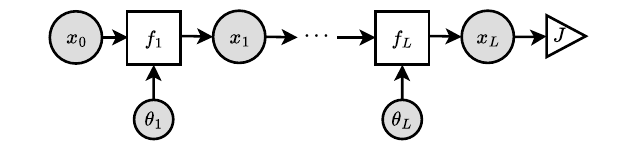} 
            \caption{Backprop: Forward pass (Phase I)}
            \label{fig:bp_fp}
        \end{subfigure}
        \\[0.3cm]
        \begin{subfigure}[b]{0.48\textwidth}
            \centering
            \includegraphics[width=\textwidth]{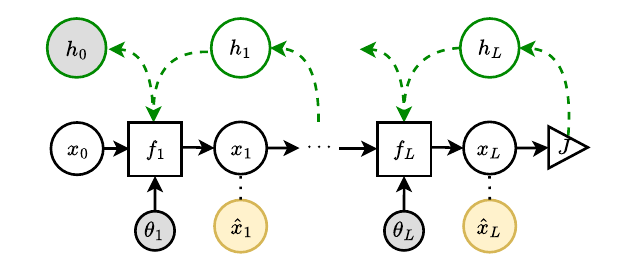}
            \caption{Moonwalk: Computing $h_0$ (Phase II)}
            \label{fig:h0_bp}
        \end{subfigure}
        &
        \begin{subfigure}[b]{0.48\textwidth}
            \centering
            \includegraphics[width=\textwidth]{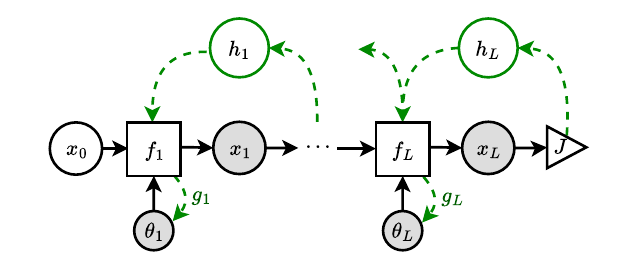} 
            \caption{Backprop: Computing gradients (Phase II)}
            \label{fig:bp_bp}
        \end{subfigure}
        \\[0.3cm]
        \begin{subfigure}[b]{0.48\textwidth}
            \centering
            \includegraphics[width=\textwidth]{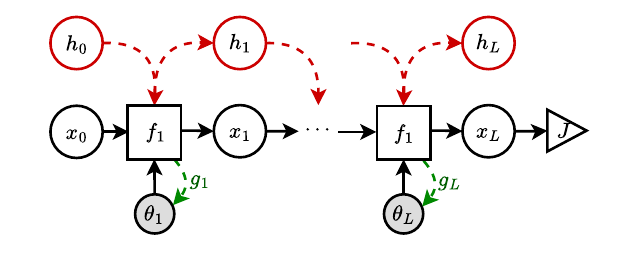}
            \caption{Moonwalk: Final gradients (Phase III)}
            \label{fig:grads_inv}
        \end{subfigure}
        &
        \begin{subfigure}[b]{0.48\textwidth}
            \centering
            \includegraphics[width=\textwidth]{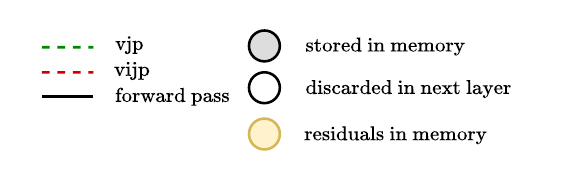}
            \vspace*{0.2cm}
            \caption*{\textbf{Legend}}
        \end{subfigure}
    \end{tabular}
    \caption{The computation flow in Moonwalk (left column) and Backprop (right column). 
    Backprop typically stores all activations from Phase I to Phase II, to facilitate the computation of both input cotangents and parameter gradients via \textit{vjp} operators.
    In contrast, Moonwalk stores only the subset $\hat{x}_i$ of residuals that is needed to compute input cotangents. }
    \label{fig:diagram}
\end{figure*}

\subsection{Notation}
\label{subsec:notation}

We consider a neural network $f_\theta: \inputspace \rightarrow \outputspace$ composed of $L$ sequential layers with parameters $\theta = \{\theta_i\}_{i=1}^L$, where $\theta_i \in \mathbb{R}^{d_i}$ denotes the parameters of layer $i$. Let $x_0 \in \inputspace$ denote the input to the network. The output of layer $i = 1, \ldots, L$ is given by
\[
x_i = f_i(x_{i-1}; \theta_i) \in \mathbb{R}^{n_i},
\]
where $n_i$ is the dimensionality of the layer's output. Let $J_\theta(x_0) = J(f_\theta(x_0))$ be a scalar-valued loss function. Our goal is to compute the gradient $\nabla_\theta J_\theta$ for use in gradient-based optimization.

In convolutional layers, we distinguish between abstract vector notation (as above) and structured tensor notation. Specifically, we represent the input to a convolutional layer as a tensor $x \in \mathbb{R}^{\mathbf{n} \times m}$, where $\mathbf{n} \in \mathbb{N}^d$ denotes the spatial shape and $m$ the number of input channels. Bold symbols $\mathbf{i}, \mathbf{k}, \mathbf{p}, \mathbf{s}$ denote multi-dimensional spatial values (respectively, element coordinates, kernel sizes, padding, and stride), while $c, c'$ index input and output channels, respectively. Outputs of convolutional layers are similarly denoted $x' \in \mathbb{R}^{\mathbf{n}' \times m'}$.

Throughout, we refer to the Jacobian–vector product, vector–Jacobian product, and vector–inverse-Jacobian product as \textit{jvp}, \textit{vjp}, and \textit{vijp}, respectively:
\begin{align}
    \text{jvp}(f, x, u) &= \left(\nicefrac{\partial f}{\partial x}\right) u, \\
    \text{vjp}(f, x, v) &= v \left(\nicefrac{\partial f}{\partial x}\right), \\
    \text{vijp}(f, x, v) &= v \left(\nicefrac{\partial f}{\partial x}\right)^+,
\end{align}
and similarly with respect to the parameters $\theta$ in place of the input $x$, where above $u$ is a tangent column vector, $v$ is a cotangent row vector, and $(\cdot)^+$ denotes any right-inverse of the Jacobian matrix. While \textit{jvp} and \textit{vjp} are standard primitives in autodiff frameworks such as JAX~\citep{jax2018github}, \textit{vijp} is a custom operator introduced in this work (see Appendix ~\ref{app:vijp_conv2d}).

\subsection{Preliminaries}
The parameter gradient of layer $i = 1, \ldots, L$ can be written using the chain rule as
\begin{align}
    \frac{\partial J}{\partial \theta_i} = \frac{\partial J}{\partial x_L} \left( \prod_{j=L}^{i+1}\frac{\partial x_j}{\partial x_{j-1}} \right) \frac{\partial x_i}{\partial \theta _i}. \label{eq:naive-forward}
\end{align}
While Backprop computes the product in \Cref{eq:naive-forward} from left to right, forward-mode differentiation is an alternative approach that computes it from right to left.
The suffix of the product can be computed during the forward execution of the function, such that, unlike in Backprop, no residuals need to be stored. On the other hand, while the prefixes are vectors of dimension $n_j$ that can be computed using \textit{vjp}, the suffixes are matrices of dimensions $n_j \times d_i$. 
To avoid storing these matrices in memory, they are commonly computed column-by-column using \textit{jvp}, however this method's time complexity is significantly larger than Backprop's~(see \Cref{tab:orders}).

\begin{defn}[Submersion]
\label{def:submersion}
A differentiable function $f: \inputspace \rightarrow \outputspace$ is a \emph{submersion} if its differential $\mathrm{d} f(x)$ is surjective for all $x \in \inputspace$.
\end{defn}

This general definition from differential topology implies, in the special case of real vector spaces, that the Jacobian $\nicefrac{\partial f}{\partial x} \in \mathbb{R}^{n' \times n}$ is right-invertible, which requires $n' \le n$ and full row rank for all $x$. We call a neural network \emph{submersive} if each of its layers is a submersion with respect to its input, for all valid parameters. 
Note that all invertible networks are submersive, but the converse is not true. 
We also emphasize that the surjectivity of a Jacobian implies that the layer has a trivial cokernel. In a case when a Jacobian is not surjective, in order to compute \textit{vijp}, we need to store additional information that is required to reconstruct the cotangent~(see \cref{sec:frag}).

\section{Moonwalk}

\subsection{Overview}

Moonwalk, in its main variant that mixes the forward and reverse modes, executes gradient computation in three phases~(\cref{fig:diagram}):

\begin{description}
    \item[Phase I: Memory-efficient forward pass.]
    
    In the first phase, Moonwalk performs a forward pass while storing only a subset of residuals, specifically those required to compute the loss gradients (cotangents) with respect to the input of each layer.
    
    For commonly used layer types, such as dense and convolutional layers, this subset of stored values is substantially smaller than the complete set of residuals required to also compute gradients with respect to the parameters. For non-submersive layers, we additionally store minimal information required to reconstruct cotangents~(\cref{sec:frag}). In practice, this phase requires about $2-3\times$ less memory than the first phase of Backprop.
    
    \item[Phase II: Input cotangent recovery.]
    The second phase is very similar to the second phase of Backprop and does not add any memory overhead beyond what was stored in Phase I. Cotangent vectors with respect to the input are computed by propagating the loss gradients backwards through the network, similar to traditional backpropagation, but retracing only the computation path from the loss back to the input and ignoring gradients with respect to parameters.
    
    \item[Phase III: Parameter gradients via Forward.]
    We sequentially traverse the network in the forward direction, computing each layer's activation together with its \textbf{output cotangent} and the gradients with respect to the layer's \textbf{parameters}.
    
    The output cotangent is obtained from the layer's \textbf{input cotangent} using our custom operator, the \textbf{vector–inverse-Jacobian product} (\textit{vijp}), which efficiently applies the right-inverse of the layer's Jacobian without materializing the full matrix. When the Jacobian has a non-trivial cokernel and lacks a right-inverse, we reconstruct the output cotangent using stored information from \textbf{Phase I}~(see \cref{sec:frag}). Finally, the gradient with respect to the layer's \textbf{parameters} is computed via a standard vector–Jacobian product (\textit{vjp}).

\end{description}

\textbf{Pure-forward} Moonwalk differs from the above mixed-mode variant by replacing Phases I and II with forward-mode computation of the input cotangent, avoiding any residual storage at the cost of increased computation.

Regardless of how the input cotangent is obtained, Moonwalk needs nothing more for its Phase III in submersive networks, making the memory overhead of this phase constant in the network depth.
Mixed-mode Moonwalk can also be applied to non-submersive layers through \textbf{gradient checkpointing} in Phase II, where only the cotangents required for \textit{vjp} with respect to the parameters are checkpointed, avoiding full residual storage. This strategy can be further refined into \textbf{fragmental cotangent checkpointing}, a memory-efficient variant that stores minimal cotangent fragments, enabling parallel reconstruction in Phase III~(\cref{sec:frag}).

Because submersive networks are a much larger superset of invertible networks, Moonwalk is usable in some architectures where inversion-based methods are not, such as networks that reduce the dimensionality of their input or have non-injective activation functions.

Moonwalk can also be combined with checkpointing, also known as rematerialization, to reduce the effective network depth in Phase II.

\subsection{The Moonwalk Identity}

In order to benefit from the memory advantage of forward-mode gradient computation, while keeping the time complexity similar to that of Backprop and avoiding the introduction of noisy gradients through projection, we first restrict our attention to the class of submersive networks, 
in which the Jacobian of each layer with respect to its input is guaranteed to be right-invertible (see Definition \ref{def:submersion}). 
Then we can rewrite layer $i$'s parameter gradient of the loss, for $i = 1, \ldots, L$, as
\begin{align}
g_i  \coloneqq{}& \frac{\partial J}{\partial \theta_i} = \frac{\partial J}{\partial x_i}\frac{\partial x_i}{\partial \theta_i} = \frac{\partial J}{\partial x_i} \frac{\partial x_i}{\partial x_0} \left(\frac{\partial x_i}{\partial x_0}\right)^+ \frac{\partial x_i}{\partial \theta_i} \label{eq:g_i_def}\\
={}& \frac{\partial J}{\partial x_0}\left(\prod_{j=i}^1\frac{\partial x_j}{\partial x_{j-1}}\right)^+ \frac{\partial x_i}{\partial \theta_i} \label{eq:g_i_def_2}\\
={}& \frac{\partial J}{\partial x_0}\prod_{j=1}^i\left(\frac{\partial x_j}{\partial x_{j-1}}\right)^+ \frac{\partial x_i}{\partial \theta_i}. \label{eq:ifp-main}
\end{align}
Given the input cotangent $\frac{\partial J}{\partial x_0}$, we can recurse on prefixes of \Cref{eq:ifp-main} to compute each layer's parameter gradient in forward-mode. To this end, denote the input cotagent of layer $i = 1, \ldots, L$ by
\begin{align}
    h_i \coloneqq \frac{\partial J}{\partial x_i} = \frac{\partial J}{\partial x_0} \prod_{j=1}^i\left(
    \frac{\partial x_j}{\partial x_{j-1}}\right)^+,
\end{align}
and the network's input cotangent by $h_0 \coloneqq \frac{\partial J}{\partial x_0}$. This formulation allows us to compute each cotangent $h_i$ and parameter gradient $g_i$, as we evaluate $f_i(x_{i-1}; \theta_i)$ in a forward pass, from only the layer input cotangent $h_{i-1}$ and the local Jacobians, without any residuals:
\begin{align}
    h_i = h_{i-1} \left(\frac{\partial x_i}{\partial x_{i-1}}\right)^+ = \text{vijp}(f_i, x_{i-1}, h_{i-1}),  \label{eq:h_i}
\end{align}
and from \Cref{eq:ifp-main} we have
\begin{align}
    g_i = h_{i} \frac{\partial x_i}{\partial \theta_i} = \text{vjp}(f_i, \theta_i, h_i). \label{eq:g_i}
\end{align}
Assuming that we have the input cotangent $h_0$, we can construct the parameter gradient for each layer on-the-fly by the two operators in \Cref{eq:h_i,eq:g_i} and store only $h_i \in \Z{R}^{n_i}$ temporarily for the next layer's computation.
Despite the general existence of multiple right-inverses, the result is unique in submersive networks because each input cotangent is in the row-space of its layer's Jacobian.
The complete procedure is given in \cref{alg:invertible-fp} and illustrated in \cref{fig:grads_inv}.


\begin{algorithm}[t]
\caption{Moonwalk}
\label{alg:invertible-fp}
\begin{algorithmic}
    \FOR{\textbf{each} gradient step with input $x_0$}
        \STATE Compute $h_0 \gets \frac{\partial J}{\partial x_0}$
        \FOR{$i = 1, \ldots, L$}
            \STATE $x_i \gets f_i(x_{i-1}; \theta_i)$
            \STATE $h_{i} \gets \text{vijp}(f_i, x_{i-1}, h_{i-1})$
            \STATE $g_i \gets \text{vjp}(f_i, \theta_i, h_i)$
            \STATE Apply $g_i$ to $\theta_i$
        \ENDFOR
    \ENDFOR
\end{algorithmic}
\end{algorithm}

In the remainder of this section we describe the computation of $h_0$ in either reverse- or forward-mode, discuss the trade-offs of each variant, and introduce techniques offering additional improvements.
\subsection{Mixed-Mode Moonwalk} \label{sec:mixed-forward}

A key component of Moonwalk is the computation of the input cotangent $h_0$. In the mixed-mode variant, we obtain $h_0$ using reverse-mode differentiation. At first glance, this may appear to negate the memory benefits of our approach. However, the crucial observation is that computing $h_0$ only requires traversing the portion of the computation graph that contributes to the input cotangent, rather than the full set of parameter-dependent paths.

In particular, for many layers—such as convolutions—the input cotangent depends only on the layer parameters (e.g., kernel weights) and the output cotangent, but not on the full set of stored activations required to compute parameter gradients. As a result, reverse-mode computation of $h_0$ can be carried out without storing the residuals needed for parameter updates. This decoupling allows Moonwalk to retain the efficiency of reverse-mode for computing $h_0$, while avoiding its primary memory bottleneck.

Once $h_0$ is obtained, all parameter gradients are recovered in a single forward sweep using the \textit{vijp} operator, eliminating the need to store intermediate activations entirely in submersive networks. This hybrid strategy combines the strengths of reverse- and forward-mode differentiation: reverse-mode efficiently computes a single cotangent, while forward-mode reconstructs all remaining gradients with reduced memory overhead.

From a systems perspective, this design also contrasts with standard backpropagation, where parameter gradients are typically accumulated across the entire network before any updates are applied. In Moonwalk, parameter gradients are computed sequentially during the forward sweep and need not be stored simultaneously, further reducing peak memory usage. This distinction is particularly impactful in architectures with large parameter counts, where gradient storage can dominate memory consumption.

Finally, we note that alternative choices for the reconstruction “seed” can be used in place of $h_0$, such as the first-layer parameter gradient $g_0$ or its output cotangent $h_1$, leading to modest efficiency gains in some architectures. These variants highlight the flexibility of the mixed-mode formulation and promise further opportunities for optimizing memory–compute trade-offs.
\subsection{Pure-Forward Moonwalk}\label{sec:pure-forward}

An alternative way to obtain the input cotangent $h_0$ is to compute it entirely in forward-mode. In this setting, each component $h_{0,j}$ is obtained by propagating a standard basis vector $e_j$ through the network using Jacobian–vector products (\textit{jvp}). Concretely, this corresponds to the recursion
\[
\frac{\partial x_i}{\partial x_0} e_j = \mathrm{jvp}(f_i, x_{i-1}, \tfrac{\partial x_{i-1}}{\partial x_0} e_j),
\]
evaluated independently for each input dimension $h_{0, j}$. 

This approach eliminates the need for reverse-mode entirely and requires no storage of intermediate activations. Instead, only the components of $h_0$ are retained, making the method maximally memory-efficient. As in the mixed-mode variant, once $h_0$ is computed, all parameter gradients are recovered in a single forward sweep via the \textit{vijp} operator.

However, this benefit comes at a computational cost: constructing $h_0$ requires one forward-mode pass per input dimension. As a result, the method scales linearly with the input size and can become prohibitively expensive in high-dimensional settings. Consequently, pure-forward Moonwalk is most suitable when the input dimension is small or when memory constraints dominate compute considerations. In such regimes, it provides a simple and fully forward-mode alternative that highlights the conceptual generality of the Moonwalk framework.
\subsection{Residual Impact}

A central advantage of Moonwalk is its reduced reliance on storing intermediate residuals. In contrast to Backprop, which retains all activations required to compute vector–Jacobian products, Moonwalk stores only a subset sufficient for backpropagating and reconstructing input/output cotangents. As illustrated in~\cref{fig:h0_fp}, this corresponds to storing compressed representations $\hat{x}_i$ rather than full activations.

This reduction can be substantial in certain common architectures. For example, in a sequence of convolutional layers with LeakyReLU activations, Backprop must store the full input to each convolution to compute parameter gradients. In contrast, Moonwalk only requires storing the sign of each activation to evaluate the LeakyReLU \textit{vjp}, which dramatically reduces memory usage. In practice, this representation can be up to $16$--$32\times$ smaller than storing full-precision activations.

More generally, 
this shift from activation storage to compact structural summaries enables significant memory savings over Backprop, particularly in networks with large feature maps or high-resolution inputs. As a result, Moonwalk can scale to deeper architectures or larger batch sizes under the same memory budget.

\section{Submersive Convolutional Layers}

Convolutional layers are structured linear operators that can use relatively few parameters to transform large inputs while respecting (typically translational) symmetries.
As such, they are ideal for the performance gains offered by Moonwalk, both because Backprop residuals are large and because the \textit{vijp} operator can leverage the convolutional structure.
The following lemma provides sufficient conditions for a convolutional layer to be submersive.

\begin{lemma}[Sufficient Conditions for Submersive Convolutional Layers]
\label{lem:submersive_convolution}
A channel-wise convolution with input $x \in \mathbb{R}^{\mathbf{n}\times m}$, kernel $w \in \mathbb{R}^{\mathbf{k}\times m \times m'}$, output $x' \in \mathbb{R}^{\mathbf{n}'\times m'}$, padding $\mathbf{p}$, and stride $\mathbf{s}$, defined by
\eqn{
x'_{\mathbf{i}', c'} = \sum_{\mathbf{j}, c} w_{\mathbf{j}, c, c'} \cdot x_{\mathbf{s}\mathbf{i}' + \mathbf{j} - \mathbf{p}, c},
}
is submersive—i.e., its input–output Jacobian is right-invertible—if the following conditions hold:
\begin{itemize}[nosep]
    \item[\textnormal{(i)}] \textbf{Spatial bounds:} $\mathbf{k} > \mathbf{p}$, $\mathbf{s} > \mathbf{p}$, and $\mathbf{n} > \mathbf{s}(\mathbf{n}' - 1)$;
    \item[\textnormal{(ii)}] \textbf{Channel-wise triangularity:} $w_{\mathbf{p}, c, c'} = 0$ for all $c < c'$ (implying $m' \le m$); and
    \item[\textnormal{(iii)}] \textbf{Diagonal support:} $w_{\mathbf{p}, c', c'} \neq 0$ for all $c' \le m'$.
\end{itemize}
\end{lemma}




 \textit{Proof outline.}
We provide a constructive proof by uniquely recovering the output cotangent of the convolutional layer from its input cotangent.
This construction will then also serve as an efficient implementation of the \textit{vijp} operator for submersive convolutional layers.

In reverse-mode, the input cotangent $h$ can be computed from the output cotangent $h'$ as:

\begin{align}
h &= \mathrm{TransposeConv}(h', w, \mathbf{n}, \mathbf{p}, \mathbf{s}) \notag \\
  &= \mathrm{Conv}(\bar{h}', w^\top, \mathbf{k} - \mathbf{p} - \mathbf{1}, \mathbf{1}) \\
h_{\mathbf{i}, c} 
  &= \sum_{\mathbf{j}, c'} w_{\mathbf{j}, c, c'} 
     \cdot \bar{h}'_{\mathbf{i} + \mathbf{p} - \mathbf{j}, c'},
     \label{eq:conv-vjp}
\end{align}

where $\bar{h}'$ is obtained from $h'$ by dilating it by $\mathbf{s}$ (i.e., $\bar{h}'_{\mathbf{s} \mathbf{i}', c'} = h'_{\mathbf{i}', c'}$ and zero elsewhere), and padding the end of each spatial dimension with $\left\{ \frac{\mathbf{n} + 2\mathbf{p} - \mathbf{k}}{\mathbf{s}} \right\} \mathbf{s}$ zeros, where $\left\{\cdot\right\}$ denotes the fractional part. Kernel transposition $w^\top$ is a reversal of kernel elements along spatial dimensions and a transposition of channel dimensions.

When the convolution is submersive, the \textit{vjp} operator defined in \eqref{eq:conv-vjp} can be uniquely inverted to recover $h'$ from $h$.
The sufficient conditions in the lemma are aimed at simplifying this \textit{vijp} operator as Gaussian elimination with fixed indices.
Specifically, we prove in Appendix ~\ref{app:proof_sub} that each $h'_{\mathbf{i}', c'}$ is the leading element on the right-hand side of \eqref{eq:conv-vjp} for $h_{\mathbf{s} \mathbf{i}', c'}$, with coefficient $w_{\mathbf{p}, c', c'} \neq 0$.
It follows that
\begin{align}
&w_{\mathbf{p}, c', c'} h'_{\mathbf{i}', c'} 
   = \notag \\ & \quad h_{\mathbf{s} \mathbf{i}', c'} - \mathrm{TransposeConv}(h', \tilde{w}, 
         \mathbf{n}, \mathbf{p}, \mathbf{s})_{\mathbf{s} \mathbf{i}', c'},
\end{align}

where $\tilde{w}$ is identical to $w$ except that $\tilde{w}_{\mathbf{p}, c', c'} = 0$ for all $c'$. Because all nonzero $h'$ terms needed to compute $h'_{\mathbf{i}', c'}$ have indices smaller than $(\mathbf{i}', c')$, we can parallelize the computation over all entries with the same total index sum $t = \sum \mathbf{i}' + c'$. Pseudocode for a highly efficient parallel implementation is provided in Appendix ~\ref{app:vijp_conv2d}. This implementation enables the simultaneous computation of all spatial cotangent values, offering significant parallelization benefits. 

\subsection{Fragmental Gradient Checkpointing}\label{sec:frag}

To increase the flexibility of supported architectures, we aim to incorporate non-submersive layers. The challenge, however, is that the output cotangent of such a layer may not be uniquely recoverable from its input cotangent, while storing the full output cotangent 
can be memory-intensive. To address this, we propose 
fragmental checkpointing: instead of retaining the entire output cotangent, we selectively store a minimal subset of its elements that is sufficient to reconstruct the remaining elements through recursive elimination.

As an example, in a channel-last 1D convolution with kernel weights $\mathbf{w} \in \mathbb{R}^{k \times m \times m'}$, $p=1$, and $s=1$, an output cotangent element $h'_{i', c'}$ can be computed as
%
\begin{align}
\label{final_frag2}
h'_{i',c'} 
  &= h_{i'-1,c'} \notag \\
  &\quad - \sum_{c > c'} w_{0, c, c'} \cdot h'_{i', c'} \notag \\
  &\quad - \sum_{\substack{j \ge 1 \\ c \ge c'}} 
        w_{j, c, c'} \cdot h'_{i' - j, c'}
\end{align}
assuming that
\begin{itemize}
\item $w_{0,c,c'} = 0$ for all $c < c'$; and
\item $w_{0,c',c'} = 1$.
\end{itemize}
%
Recovering all channels $h'_{i',\cdot}$ at index $i'$ therefore requires, in addition to the input cotangent elements $h_{i'-1, \cdot}$, the previous $k-1$ output elements $h'_{i'-j, \cdot}$ (in all channels).
The full derivation is in Appendix ~\ref{app:fragmental_gradient}.


More generally, fragmental checkpointing can be applied in blocks. For block size $B$, storing just the first $k - 1$ entries per block allows reconstruction of the rest of the block. For example, given a residual tensor shaped $1024 \times 64$ (time steps $\times$ channels), with $k = 3$ and $B = 4$, only half of each block (2 elements) needs to be stored, reducing memory from $1024 \times 64$ to $512 \times 64$. Increasing the block size to $B = 16$ with the same $k$ reduces the memory cost to $\nicefrac{1}{8}$ of full checkpointing. Additional implementation details are in Appendix ~\ref{app:frag_vijp_1d}.

\Cref{tab:orders} summarizes the order of growth of time and memory in the methods we compare, and the next section evaluates them empirically. 

\section{Experiments}
We empirically evaluate the memory and computational performance of Moonwalk against Backprop on deep residual convolutional networks. 
Our setup is essentially identical to the original Resnet \cite{resnet} on Imagenet \cite{imagenet}. Our experiments are designed to isolate the memory–computation tradeoffs introduced by Moonwalk in both fully submersive networks and under fragmental checkpointing.
We also showcase that our implemented convolutional \textit{vijp} operator does not introduce a computational overhead while leading to significant memory savings. 

\subsection{Experimental Setup}
All experiments are conducted on a single NVIDIA RTX 3090 GPU (24 GB) using JAX with \texttt{jaxlib} version 0.4.30 and CUDA 12.6. We measure peak GPU memory usage using \texttt{jax.device.memory\_stats()}, and we report total wall-clock execution time for a single forward and backward pass on one batch (batch size 128), after the initial JIT compilation. We compare Moonwalk against Backprop and Backprop with rematerialization. 

\subsection{Fully Parallel Submersive 2D CNN}

We begin with a fully submersive 2D convolutional architecture. We simulate a typical setup of the most common residual CNNs architectures trained on Imagenet \cite{imagenet}. An input of size $256 \times 256 \times 3$ is first upsampled to $256 \times 256 \times 128$ channels. Each subsequent layer applies a $3 \times 3 \times 128 \times 128$ convolution with stride 2 and padding 1, allowing a fully parallelizable \textit{vijp} operator and halving the spatial resolution. All activation functions are LeakyReLU, and the final layer performs max pooling and projects the feature map to a scalar.

\begin{figure}[t]
  \centering
  \begin{subfigure}[t]{0.49\textwidth}
    \includegraphics[width=\linewidth]{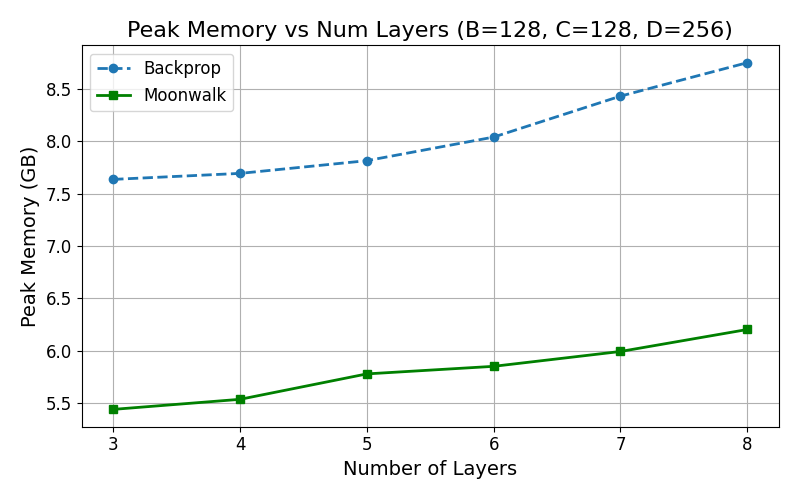}
    \caption{2D - GPU memory vs. \# of layers.}
    \label{fig:mem2d}
  \end{subfigure}
  \hfill
  \begin{subfigure}[t]{0.49\textwidth}
    \includegraphics[width=\linewidth]{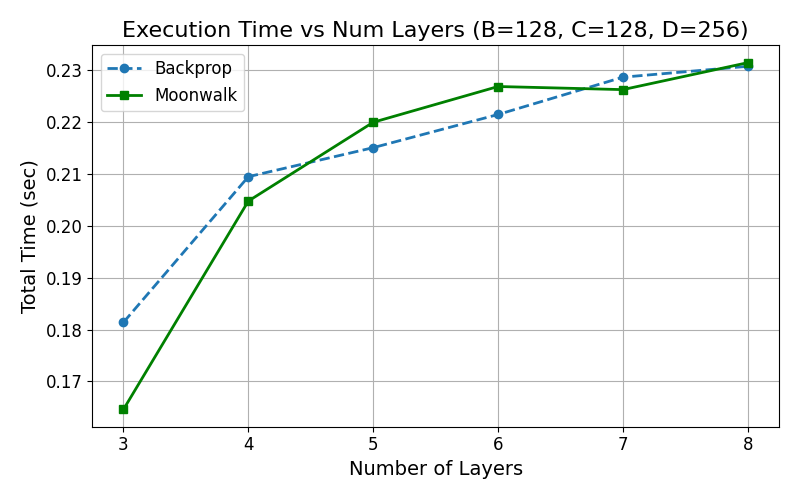}
    \caption{2D - Total time vs. \# of layers.}
    \label{fig:time2d}
  \end{subfigure}

  \caption{  
    Comparison of Backprop and Moonwalk as network depth increases in a 2D CNN.  
    (a) Moonwalk reduces peak memory by about 30\% compared to Backprop.  
    (b) Both methods incur similar runtimes, demonstrating that Moonwalk’s memory savings come at no extra compute cost.  
  }
  \label{fig:2d-performance}
\end{figure}

This configuration enables efficient memory usage, as illustrated in \cref {fig:mem2d}. Specifically, Moonwalk reduces the memory footprint from 9.5 GB to 6.6 GB in the 8-block configuration, while maintaining comparable computational performance as illustrated in \cref {fig:time2d}. This corresponds to a $30\%$ reduction in memory without sacrificing throughput. These savings are primarily due to the fully parallelizable \textit{vijp} operator, which computes all intermediate values simultaneously without sequential dependency.

\subsection{Fragmental Checkpointing}

To extend Moonwalk to non-submersive layers, we introduce \emph{fragmental gradient checkpointing}, which leverages the structure of convolutional operators. 

For implementation simplicity, we illustrate this approach on 1D convolutional networks, but note that it is equally applicable to higher spatial dimensions. In non-submersive layers, the Jacobian has a non-trivial cokernel, preventing exact recovery of cotangents through \textit{vijp} alone. To address this, we store a minimal subset of cotangent fragments during the backward pass, which are later used to reconstruct the missing gradient components during the forward sweep.

Our experimental setup mirrors the 2D case, but preserves spatial resolution at each layer by setting the stride and padding to 1. The $2048 \times 3$ input is upsampled to $2048 \times 256$, and kept at this shape across layers.
While this configuration breaks submersivity, fragmental checkpointing enables exact cotangent recovery.

\begin{figure}[t]
    \centering
    \begin{subfigure}[t]{0.49\textwidth}
        \centering
        \includegraphics[width=\linewidth]{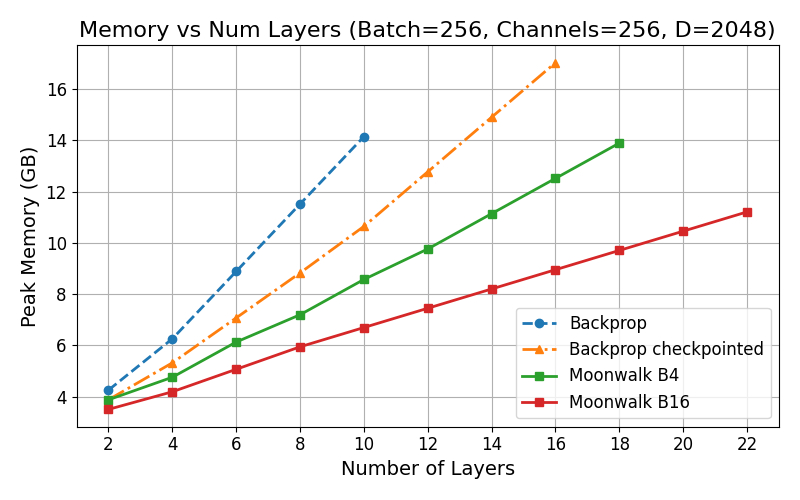}
        \caption{1D Fragmental - GPU memory vs. \# of layers.}
        \label{fig:mem1d}
    \end{subfigure}
    \hfill
    \begin{subfigure}[t]{0.49\textwidth}
        \centering
        \includegraphics[width=\linewidth]{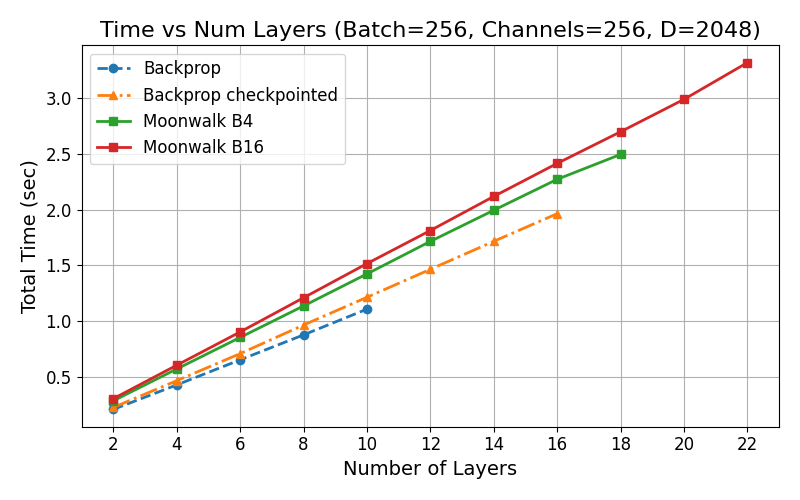}
        \caption{1D Fragmental - Total time vs. \# of layers.}
        \label{fig:time1d}
    \end{subfigure}

    \caption{
    Evaluation of Moonwalk with fragmental checkpointing on a 1D CNN.  
    (a) At fixed block size $B{=}4$, Moonwalk reduces memory usage by up to 50\% compared to Backprop.  
    (b) Varying the block size reveals a trade-off: bigger blocks require more recomputation and increase runtime.
    }
    \label{fig:1d-fragmented}
\end{figure}

This approach provides a flexible memory–compute trade-off. With block size 4—where half of the cotangent components are stored and the rest recomputed in parallel—memory usage is reduced from 14\,GB to 7\,GB ~(\cref{fig:mem1d}), a 50\% reduction. Increasing the block size further decreases memory usage, at the cost of recomputing a larger number of unstored elements, leading to higher runtime (\cref{fig:time1d}).

Overall, fragmental checkpointing enables Moonwalk to scale to deeper networks under tighter memory budgets. In our experiments, standard backpropagation fails beyond 10 layers and checkpointed backpropagation reaches up to 16 layers, whereas Moonwalk with block size 16 successfully trains networks up to 22 layers.
\subsection{Constrained Convolutions}

To enable efficient computation of the \textit{vijp}, we parametrize convolutional layers according to the submersivity conditions in Lemma~\ref{lem:submersive_convolution}. Concretely, this imposes a structured constraint on the convolutional weights—e.g., an upper-triangular channel-wise form—that guarantees the existence of efficient right-inverse Jacobian operations.

\begin{figure}[hb!]
    \centering
    \includegraphics[width=1.0\linewidth]{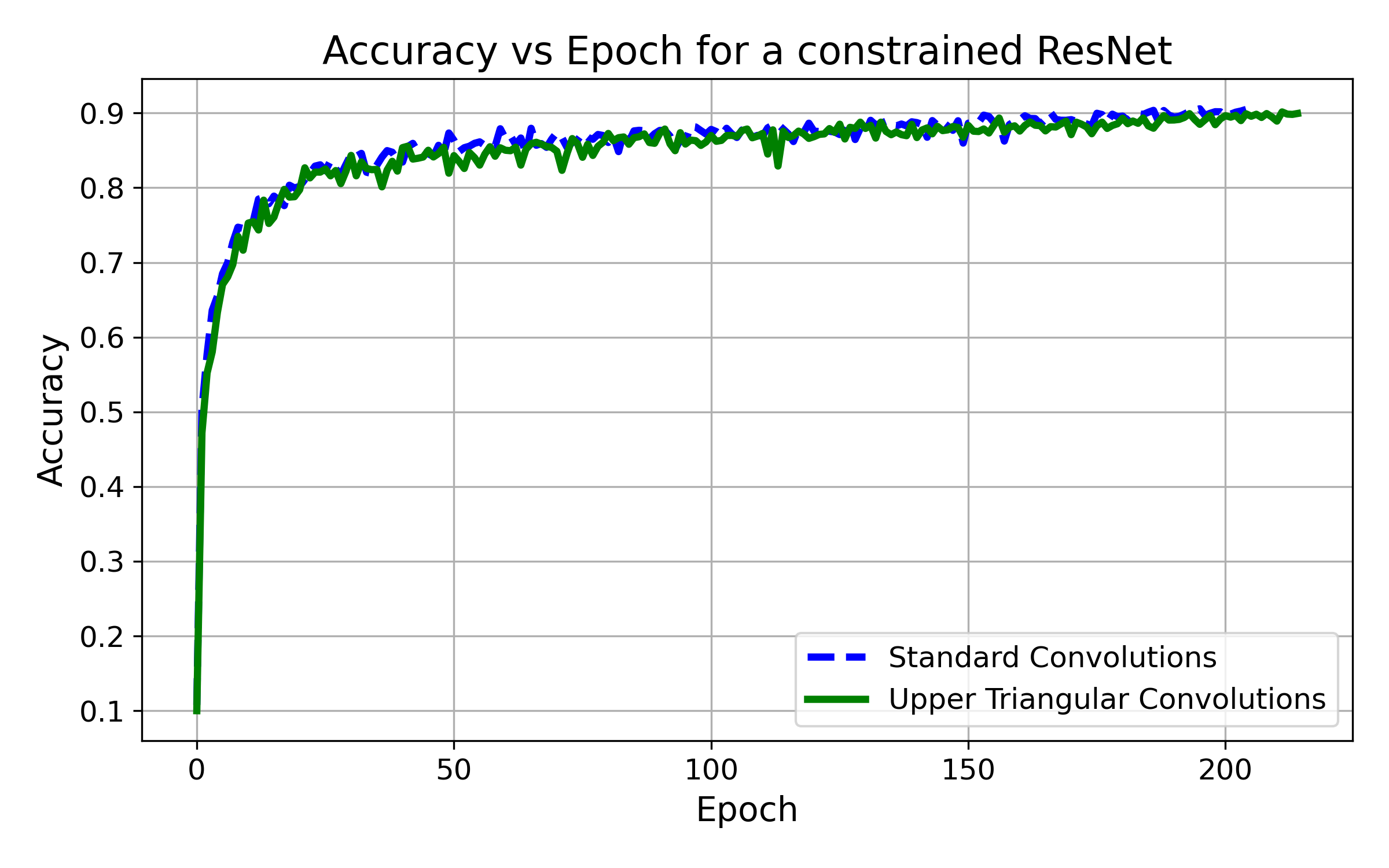}
    \caption{
    Accuracy comparison between upper-triangular convolutions (green) and standard convolutions~(blue). Both  setups converge to $\sim$90\% accuracy.
    }
    \label{fig:resnet_constrained}
\end{figure}

A natural concern is whether such constraints limit model expressivity. To evaluate this, we compare constrained and unconstrained models on a 10-layer 2D residual network, where all convolutions in the constrained variant follow the submersive parameterization. Despite the imposed structure, both models achieve comparable performance, reaching approximately 90\% test accuracy (\cref{fig:resnet_constrained}).

These results suggest that the proposed constraints do not significantly degrade representational power in practice, while enabling the efficient \textit{vijp} computations required by Moonwalk. This highlights a favorable trade-off between architectural structure and computational efficiency, and supports the viability of submersive parameterizations in realistic settings.

\section{Limitations}
\label{sec:limit}
The primary limitation of our approach is the need to implement the \textit{vijp} operator manually for each layer class, which is more involved than relying on the automatically generated \textit{vjp} operators provided by modern AD frameworks. In cases where constructing an efficient \textit{vijp} is difficult, this overhead can be mitigated through partial checkpointing of cotangents.

Despite this requirement, we find that for many practical architectures—particularly convolutional networks—efficient \textit{vijp} implementations are attainable, yielding favorable memory–compute trade-offs while remaining compatible with standard model designs.

\section{Conclusion and Future Work}
We introduced \textbf{Moonwalk}, a memory-efficient gradient computation framework with two variants—pure-forward and mixed-mode—that, in some architectures, achieves up to $\times$2 lower memory usage with comparable runtime to backpropagation. By eliminating the need to store full residuals, Moonwalk addresses a central limitation of reverse-mode AD, making it well-suited for memory-constrained settings.

We derived sufficient conditions for convolutional layers to be submersive and used this structure to design an efficient, parallelizable \textit{vijp} operator with no additional computational overhead. These results demonstrate that exact forward-mode gradient computation can be practical in dense architectures when paired with appropriate structural assumptions.

While our analysis focuses on submersive layers, Moonwalk naturally extends to general architectures via partial cotangent storage. Our fragmental gradient checkpointing further reduces memory by enabling selective storage and parallel reconstruction of gradients, highlighting a broader design space that blends forward- and reverse-mode differentiation.

Future work includes extending Moonwalk to attention-based architectures and integrating it more tightly into modern deep learning frameworks.
\section*{Acknowledgements}
Authors Krylov and Karamzade were supported by
Hasso Plattner Foundation Fellowships. This work was
funded in part by the National Science Foundation (Award \#2321786). 

\bibliographystyle{plainnat}
\bibliography{main}

@inproceedings{jacobsen2018revnet,
  title={i-RevNet: Deep Invertible Networks},
  author={Jacobsen, J{\"o}rn-Henrik and Smeulders, Arnold WM and Oyallon, Edouard},
  booktitle={International Conference on Learning Representations},
  year={2018}
}

@article{kingma2018glow,
  title={Glow: Generative flow with invertible 1x1 convolutions},
  author={Kingma, Durk P and Dhariwal, Prafulla},
  journal={Advances in neural information processing systems},
  volume={31},
  year={2018}
}

@misc{reversellm,
      title={Reversing Large Language Models for Efficient Training and Fine-Tuning}, 
      author={Eshed Gal and Moshe Eliasof and Javier Turek and Uri Ascher and Eran Treister and Eldad Haber},
      year={2025},
      eprint={2512.02056},
      archivePrefix={arXiv},
      primaryClass={cs.CL},
      url={https://arxiv.org/abs/2512.02056}, 
}

@inbook{inverseaccum,
   title={Automatic Differentiation: Inverse Accumulation Mode},
   ISBN={9781611979039},
   url={http://dx.doi.org/10.1137/1.9781611979039.1},
   DOI={10.1137/1.9781611979039.1},
   booktitle={Proceedings of the 2024 International Conference on Algorithmic Differentiation (AD)},
   publisher={Society for Industrial and Applied Mathematics},
   author={Pearlmutter, Barak A. and Siskind, Jeffrey Mark},
   year={2026},
   month=jan, pages={1–12} }

@article{
sharding,
title={Activation sharding for scalable training of large models},
author={Xingzi Xu and Amir Tavanaei and Kavosh Asadi and Karim Bouyarmane},
journal={Transactions on Machine Learning Research},
issn={2835-8856},
year={2025},
url={https://openreview.net/forum?id=kQCuMcEneq},
note={}
}

@misc{deq,
      title={Deep Equilibrium Models}, 
      author={Shaojie Bai and J. Zico Kolter and Vladlen Koltun},
      year={2019},
      eprint={1909.01377},
      archivePrefix={arXiv},
      primaryClass={cs.LG},
      url={https://arxiv.org/abs/1909.01377}, 
}

@misc{reversecol,
      title={Reversible Column Networks}, 
      author={Yuxuan Cai and Yizhuang Zhou and Qi Han and Jianjian Sun and Xiangwen Kong and Jun Li and Xiangyu Zhang},
      year={2023},
      eprint={2212.11696},
      archivePrefix={arXiv},
      primaryClass={cs.CV},
      url={https://arxiv.org/abs/2212.11696}, 
}

@misc{autohete,
      title={AutoHete: An Automatic and Efficient Heterogeneous Training System for LLMs}, 
      author={Zihao Zeng and Chubo Liu and Xin He and Juan Hu and Yong Jiang and Fei Huang and Kenli Li and Wei Yang Bryan Lim},
      year={2025},
      eprint={2503.01890},
      archivePrefix={arXiv},
      primaryClass={cs.LG},
      url={https://arxiv.org/abs/2503.01890}, 
}

@misc{lynx,
      title={Optimizing Large Model Training through Overlapped Activation Recomputation}, 
      author={Ping Chen and Wenjie Zhang and Shuibing He and Weijian Chen and Siling Yang and Kexin Huang and Yanlong Yin and Xuan Zhan and Yingjie Gu and Zhuwei Peng and Yi Zheng and Zhefeng Wang and Gang Chen},
      year={2025},
      eprint={2406.08756},
      archivePrefix={arXiv},
      primaryClass={cs.DC},
      url={https://arxiv.org/abs/2406.08756}, 
}

@inproceedings{mario,
author = {Liu, Weijian and Li, Mingzhen and Tan, Guangming and Jia, Weile},
title = {Mario: Near Zero-cost Activation Checkpointing in Pipeline Parallelism},
year = {2025},
isbn = {9798400714436},
publisher = {Association for Computing Machinery},
address = {New York, NY, USA},
url = {https://doi.org/10.1145/3710848.3710878},
doi = {10.1145/3710848.3710878},
booktitle = {Proceedings of the 30th ACM SIGPLAN Annual Symposium on Principles and Practice of Parallel Programming},
pages = {197–211},
numpages = {15},
location = {Las Vegas, NV, USA},
series = {PPoPP '25}
}

@InProceedings{hiremate,
  title = 	 {{H}i{R}emate: Hierarchical Approach for Efficient Re-materialization of Neural Networks},
  author =       {Gusak, Julia and Zhao, Xunyi and Le Hellard, Th\'{e}otime and Li, Zhe and Eyraud-Dubois, Lionel and Beaumont, Olivier},
  booktitle = 	 {Proceedings of the 42nd International Conference on Machine Learning},
  pages = 	 {21418--21443},
  year = 	 {2025},
  editor = 	 {Singh, Aarti and Fazel, Maryam and Hsu, Daniel and Lacoste-Julien, Simon and Berkenkamp, Felix and Maharaj, Tegan and Wagstaff, Kiri and Zhu, Jerry},
  volume = 	 {267},
  series = 	 {Proceedings of Machine Learning Research},
  month = 	 {13--19 Jul},
  publisher =    {PMLR},
  url = 	 {https://proceedings.mlr.press/v267/gusak25a.html}
}

@misc{rear,
      title={Reducing Activation Recomputation in Large Transformer Models}, 
      author={Vijay Korthikanti and Jared Casper and Sangkug Lym and Lawrence McAfee and Michael Andersch and Mohammad Shoeybi and Bryan Catanzaro},
      year={2022},
      eprint={2205.05198},
      archivePrefix={arXiv},
      primaryClass={cs.LG},
      url={https://arxiv.org/abs/2205.05198}, 
}

@article{checkmate,
  author       = {Paras Jain and
                  Ajay Jain and
                  Aniruddha Nrusimha and
                  Amir Gholami and
                  Pieter Abbeel and
                  Kurt Keutzer and
                  Ion Stoica and
                  Joseph E. Gonzalez},
  title        = {Checkmate: Breaking the Memory Wall with Optimal Tensor Rematerialization},
  journal      = {CoRR},
  volume       = {abs/1910.02653},
  year         = {2019},
  url          = {http://arxiv.org/abs/1910.02653},
  eprinttype   = {arXiv},
  eprint       = {1910.02653},
  bibsource    = {dblp computer science bibliography, https://dblp.org}
}

@misc{rockmate,
      title={Rockmate: an Efficient, Fast, Automatic and Generic Tool for Re-materialization in PyTorch}, 
      author={Xunyi Zhao and Théotime Le Hellard and Lionel Eyraud and Julia Gusak and Olivier Beaumont},
      year={2023},
      eprint={2307.01236},
      archivePrefix={arXiv},
      primaryClass={cs.LG},
      url={https://arxiv.org/abs/2307.01236}, 
}

@inproceedings{kumar,
 author = {Kumar, Ravi and Purohit, Manish and Svitkina, Zoya and Vee, Erik and Wang, Joshua},
 booktitle = {Advances in Neural Information Processing Systems},
 editor = {H. Wallach and H. Larochelle and A. Beygelzimer and F. d\textquotesingle Alch\'{e}-Buc and E. Fox and R. Garnett},
 pages = {},
 publisher = {Curran Associates, Inc.},
 title = {Efficient Rematerialization for Deep Networks},
 url = {https://proceedings.neurips.cc/paper_files/paper/2019/file/ffe10334251de1dc98339d99ae4743ba-Paper.pdf},
 volume = {32},
 year = {2019}
}

@article{mackay2018reversible,
  title={Reversible recurrent neural networks},
  author={MacKay, Matthew and Vicol, Paul and Ba, Jimmy and Grosse, Roger B},
  journal={Advances in Neural Information Processing Systems},
  volume={31},
  year={2018}
}

@inproceedings{mangalam2022reversible,
  title={Reversible vision transformers},
  author={Mangalam, Karttikeya and Fan, Haoqi and Li, Yanghao and Wu, Chao-Yuan and Xiong, Bo and Feichtenhofer, Christoph and Malik, Jitendra},
  booktitle={Proceedings of the IEEE/CVF Conference on Computer Vision and Pattern Recognition},
  pages={10830--10840},
  year={2022}
}

@inproceedings{revnet,
 author = {Gomez, Aidan N and Ren, Mengye and Urtasun, Raquel and Grosse, Roger B},
 booktitle = {Advances in Neural Information Processing Systems},
 editor = {I. Guyon and U. Von Luxburg and S. Bengio and H. Wallach and R. Fergus and S. Vishwanathan and R. Garnett},
 pages = {},
 publisher = {Curran Associates, Inc.},
 title = {The Reversible Residual Network: Backpropagation Without Storing Activations},
 url = {https://proceedings.neurips.cc/paper_files/paper/2017/file/f9be311e65d81a9ad8150a60844bb94c-Paper.pdf},
 volume = {30},
 year = {2017}
}

@inproceedings{bulo2018place,
  title={In-place activated batchnorm for memory-optimized training of dnns},
  author={Bulo, Samuel Rota and Porzi, Lorenzo and Kontschieder, Peter},
  booktitle={Proceedings of the IEEE Conference on Computer Vision and Pattern Recognition},
  pages={5639--5647},
  year={2018}
}

@article{gruslys2016memory,
  title={Memory-efficient backpropagation through time},
  author={Gruslys, Audrunas and Munos, R{\'e}mi and Danihelka, Ivo and Lanctot, Marc and Graves, Alex},
  journal={Advances in neural information processing systems},
  volume={29},
  year={2016}
}

@incollection{martens2012training,
  title={Training deep and recurrent networks with hessian-free optimization},
  author={Martens, James and Sutskever, Ilya},
  booktitle={Neural Networks: Tricks of the Trade: Second Edition},
  pages={479--535},
  year={2012},
  publisher={Springer}
}

@inproceedings{rezende2015variational,
  title={Variational inference with normalizing flows},
  author={Rezende, Danilo and Mohamed, Shakir},
  booktitle={International conference on machine learning},
  pages={1530--1538},
  year={2015},
  organization={PMLR}
}

@inproceedings{jaderberg2017decoupled,
  title={Decoupled neural interfaces using synthetic gradients},
  author={Jaderberg, Max and Czarnecki, Wojciech Marian and Osindero, Simon and Vinyals, Oriol and Graves, Alex and Silver, David and Kavukcuoglu, Koray},
  booktitle={International conference on machine learning},
  pages={1627--1635},
  year={2017},
  organization={PMLR}
}

@article{dinh2014nice,
  title={Nice: Non-linear independent components estimation},
  author={Dinh, Laurent and Krueger, David and Bengio, Yoshua},
  journal={arXiv preprint arXiv:1410.8516},
  year={2014}
}

@inproceedings{ren2022scaling,
  title={Scaling Forward Gradient With Local Losses},
  author={Ren, Mengye and Kornblith, Simon and Liao, Renjie and Hinton, Geoffrey},
  booktitle={The Eleventh International Conference on Learning Representations},
  year={2022}
}

@inproceedings{silver2021learning,
  title={Learning by Directional Gradient Descent},
  author={Silver, David and Goyal, Anirudh and Danihelka, Ivo and Hessel, Matteo and van Hasselt, Hado},
  booktitle={International Conference on Learning Representations},
  year={2021}
}

@misc{baydin2022gradients,
      title={Gradients without Backpropagation}, 
      author={Atılım Güneş Baydin and Barak A. Pearlmutter and Don Syme and Frank Wood and Philip Torr},
      year={2022},
      eprint={2202.08587},
      archivePrefix={arXiv},
      primaryClass={cs.LG}
}

@article{chen2016training,
  title={Training deep nets with sublinear memory cost},
  author={Chen, Tianqi and Xu, Bing and Zhang, Chiyuan and Guestrin, Carlos},
  journal={arXiv preprint arXiv:1604.06174},
  year={2016}
}

@software{jax2018github,
  author = {James Bradbury and Roy Frostig and Peter Hawkins and Matthew James Johnson and Chris Leary and Dougal Maclaurin and George Necula and Adam Paszke and Jake Vander{P}las and Skye Wanderman-{M}ilne and Qiao Zhang},
  title = {{JAX}: composable transformations of {P}ython+{N}um{P}y programs},
  url = {http://github.com/google/jax},
  version = {0.3.13},
  year = {2018},
}

@misc{forwardmatchback,
      title={Can Forward Gradient Match Backpropagation?}, 
      author={Louis Fournier and Stéphane Rivaud and Eugene Belilovsky and Michael Eickenberg and Edouard Oyallon},
      year={2023},
      eprint={2306.06968},
      archivePrefix={arXiv},
      primaryClass={cs.LG}
}

@InProceedings{reducing_activations,
  title = 	 {Few-bit Backward: Quantized Gradients of Activation Functions for Memory Footprint Reduction},
  author =       {Novikov, Georgii Sergeevich and Bershatsky, Daniel and Gusak, Julia and Shonenkov, Alex and Dimitrov, Denis Valerievich and Oseledets, Ivan},
  booktitle = 	 {Proceedings of the 40th International Conference on Machine Learning},
  pages = 	 {26363--26381},
  year = 	 {2023},
  editor = 	 {Krause, Andreas and Brunskill, Emma and Cho, Kyunghyun and Engelhardt, Barbara and Sabato, Sivan and Scarlett, Jonathan},
  volume = 	 {202},
  series = 	 {Proceedings of Machine Learning Research},
  month = 	 {23--29 Jul},
  publisher =    {PMLR},
  url = 	 {https://proceedings.mlr.press/v202/novikov23a.html}
}

@ARTICLE{OldForward,
  author={Williams, Ronald J. and Zipser, David},
  journal={Neural Computation}, 
  title={A Learning Algorithm for Continually Running Fully Recurrent Neural Networks}, 
  year={1989},
  volume={1},
  number={2},
  pages={270-280},
  doi={10.1162/neco.1989.1.2.270}}

@inproceedings{imagenet,
  author    = {Jia Deng and Wei Dong and Richard Socher and Li-Jia Li and Kai Li and Li Fei-Fei},
  title     = {ImageNet: A Large-Scale Hierarchical Image Database},
  booktitle = {2009 IEEE Conference on Computer Vision and Pattern Recognition (CVPR)},
  year      = {2009},
  pages     = {248--255},
  doi       = {10.1109/CVPR.2009.5206848},
  url       = {https://ieeexplore.ieee.org/document/5206848}
}

@misc{resnet,
      title={Deep Residual Learning for Image Recognition}, 
      author={Kaiming He and Xiangyu Zhang and Shaoqing Ren and Jian Sun},
      year={2015},
      eprint={1512.03385},
      archivePrefix={arXiv},
      primaryClass={cs.CV}
}

\clearpage
\section*{Checklist}

\begin{enumerate}

  \item For all models and algorithms presented, check if you include:
  \begin{enumerate}
    \item A clear description of the mathematical setting, assumptions, algorithm, and/or model. [Yes]
    \item An analysis of the properties and complexity (time, space, sample size) of any algorithm. [Yes]
    \item (Optional) Anonymized source code, with specification of all dependencies, including external libraries. [Yes / in supplemental material]
  \end{enumerate}

  \item For any theoretical claim, check if you include:
  \begin{enumerate}
    \item Statements of the full set of assumptions of all theoretical results. [Yes]
    \item Complete proofs of all theoretical results. [Yes]
    \item Clear explanations of any assumptions. [Yes]     
  \end{enumerate}

  \item For all figures and tables that present empirical results, check if you include:
  \begin{enumerate}
    \item The code, data, and instructions needed to reproduce the main experimental results (either in the supplemental material or as a URL). [Yes in supplemental]
    \item All the training details (e.g., data splits, hyperparameters, how they were chosen). [Yes]
    \item A clear definition of the specific measure or statistics and error bars (e.g., with respect to the random seed after running experiments multiple times). [Not Applicable]
    \item A description of the computing infrastructure used. (e.g., type of GPUs, internal cluster, or cloud provider). [Yes]
  \end{enumerate}

  \item If you are using existing assets (e.g., code, data, models) or curating/releasing new assets, check if you include:
  \begin{enumerate}
    \item Citations of the creator If your work uses existing assets. [Not Applicable]
    \item The license information of the assets, if applicable. [Not Applicable]
    \item New assets either in the supplemental material or as a URL, if applicable. [Not Applicable]
    \item Information about consent from data providers/curators. [Not Applicable]
    \item Discussion of sensible content if applicable, e.g., personally identifiable information or offensive content. [Not Applicable]
  \end{enumerate}

  \item If you used crowdsourcing or conducted research with human subjects, check if you include:
  \begin{enumerate}
    \item The full text of instructions given to participants and screenshots. [Not Applicable]
    \item Descriptions of potential participant risks, with links to Institutional Review Board (IRB) approvals if applicable. [Not Applicable]
    \item The estimated hourly wage paid to participants and the total amount spent on participant compensation. [Not Applicable]
  \end{enumerate}

\end{enumerate}

\section*{Appendix}

\section{Submersive Convolutional Layers Proof}
\label{app:proof_sub}

\setcounter{lemma}{0}
\begin{lemma}[Submersion Conditions for Convolutional Layers]
A channel-wise convolution with input $x \in \mathbb{R}^{\mathbf{n}\times m}$, kernel $w \in \mathbb{R}^{\mathbf{k}\times m \times m'}$, output $x' \in \mathbb{R}^{\mathbf{n}'\times m'}$, padding $\mathbf{p}$, and stride $\mathbf{s}$, defined by
\eqn{
x'_{\mathbf{i}', c'} = \sum_{\mathbf{j}, c} w_{\mathbf{j}, c, c'} \cdot x_{\mathbf{s}\mathbf{i}' + \mathbf{j} - \mathbf{p}, c},
}
is submersive—i.e., its input–output Jacobian is right-invertible—if the following conditions hold:
\begin{itemize}[nosep]
    \item[\textnormal{(i)}] \textbf{Spatial bounds:} $\mathbf{k} > \mathbf{p}$, $\mathbf{s} > \mathbf{p}$, and $\mathbf{n} > \mathbf{s}(\mathbf{n}' - 1)$;
    \item[\textnormal{(ii)}] \textbf{Channel-wise triangularity:} $w_{\mathbf{p}, c, c'} = 0$ for all $c < c'$ (implying $m' \le m$); and
    \item[\textnormal{(iii)}] \textbf{Diagonal support:} $w_{\mathbf{p}, c', c'} \neq 0$ for all $c' \le m'$.
\end{itemize}
\end{lemma}

\begin{proof}
As outlined in the main text, reverse-mode computes the input cotangent $h$ from the output cotangent $h'$ as

\begin{align}
h &= \mathrm{TransposeConv}(h', w, \mathbf{n}, \mathbf{p}, \mathbf{s}) \notag \\ &= \mathrm{Conv}(\bar{h}', w^\top, \mathbf{k} - \mathbf{p} - \mathbf{1}, \mathbf{1}) \\
h_{\mathbf{i}, c} &= \sum_{\mathbf{j}, c'} w_{\mathbf{j}, c, c'} \cdot \bar{h}'_{\mathbf{i} + \mathbf{p} - \mathbf{j}, c'},\label{eq:conv-vjp-apx}
\end{align}
where $\bar{h}'$ is obtained from $h'$ by dilating it by $\mathbf{s}$ (i.e., $\bar{h}'_{\mathbf{s} \mathbf{i}', c'} = h'_{\mathbf{i}', c'}$ and zero elsewhere), and padding the end of each spatial dimension with $\left\{ \frac{\mathbf{n} + 2\mathbf{p} - \mathbf{k}}{\mathbf{s}} \right\} \mathbf{s}$ zeros, where $\left\{\cdot\right\}$ denotes the fractional part. The transposed kernel $w^\top$ is obtained by reversing the kernel elements along spatial dimensions and transposing the channel dimensions.

The conditions in the lemma were selected to facilitate simple and fast Gaussian elimination.
The ability to always recover $h'$ from $h = h' \left(\nicefrac{\partial f}{\partial x}\right)$ then also serves to prove that the Jacobian is right-invertible, as required.

The elimination is by lexicographic order of the index $\mathbf{i}', c'$ of $h'$, with each $h'_{\mathbf{i}', c'}$ recovered by the equation for $h_{\mathbf{s}\mathbf{i}', \mathbf{c}'}$.
It is therefore enough to show that $h'_{\mathbf{i}', c'}$ is the lexicographically largest element on the right-hand side of \eqref{eq:conv-vjp-apx} for $h_{\mathbf{s}\mathbf{i}', \mathbf{c}'}$ with non-zero coefficient.

First, for any $0 \le \mathbf{i}' < \mathbf{n}'$, the index $\mathbf{s}\mathbf{i}'$ falls within the bounds of the input size $\mathbf{n}$ due to the assumption $\mathbf{n} > \mathbf{s}(\mathbf{n}' - 1)$.
Furthermore, note that the coefficient of $h'_{\mathbf{i}', c'} = \bar{h}'_{\mathbf{s}\mathbf{i}', c'}$ on the right-hand side of $h_{\mathbf{s}\mathbf{i}', \mathbf{c}'}$ is $w_{\mathbf{p}, c', c'}$, which is within the bounds of the kernel due to $\mathbf{k} > \mathbf{p}$ and non-zero by assumption (iii).

Next, any $\bar{h}$ with any spatial index larger than $\mathbf{s}\mathbf{i}'$ is either 0 due to the dilation structure or larger by at least $\mathbf{s}$, making its corresponding $w$ coefficient smaller than the aforementioned $\mathbf{p}$ by at least $\mathbf{s}$.
However, that index falls outside the kernel matrix due to $\mathbf{s} > \mathbf{p}$.

Finally, any $\bar{h}_{\mathbf{s}\mathbf{i}', c''}$ with the same spacial index but a channel index $c''$ larger than $c'$ has coefficient $w_{\mathbf{p}, c', c''} = 0$ by assumption (ii).
\end{proof}

\section{Fragmental Gradient Check\-point\-ing}
\label{app:fragmental_gradient}

Consider a \emph{channel-wise} 1-D convolution with stride \(s=1\), padding \(p=1\), and kernel size \(k=3\).
For its reverse pass we have, for every spatial index \(i\) and output channel \(c'\),
\begin{equation}
h_{i,c'}
  \;=\;
  \sum_{j=0}^{k-1}\;\sum_{c=0}^{m-1}
        w_{j,c',c}\,
        h'_{\,i-j+1,c}.
\label{eq:vjp-1d-frag}
\end{equation}

Because \(0\le j\le2\), the \emph{largest} cotangent index that appears on the right is
\(h'_{i+1,c}\) (from \(j=0\)), followed by \(h'_{i,c}\) and \(h'_{i-1,c}\), we have
\[
h_{i,c'}
  \;=\;
  \sum_{c}
     \bigl(
        w_{0,c',c}\,h'_{\,i+1,c}
        \;+\;
        w_{1,c',c}\,h'_{\,i,c}
        \;+\;
        w_{2,c',c}\,h'_{\,i-1,c}
     \bigr).
\]

Assuming $w_{0,c',c}=0$ for $c>c'$ and $w_{0,c',c'}=1$ for $0\le c',c<m$, we can solve for the \emph{future} cotangent
\(h'_{\,i+1,c'}\) directly as
\begin{align}
\label{eq:frag_update_simple}
h'_{\,i+1,c'} =
   h_{i,c'}
   &-
   \sum_{c<c'}        w_{0,c',c}\,h'_{\,i+1,c} \notag \\
   &-
   \sum_{j=1}^{2}\;\sum_{c} w_{j,c',c}\,h'_{\,i-j+1,c}.
\end{align}

Equation~\eqref{eq:frag_update_simple} shows that
\(h'_{\,i+1,c'}\) depends only on the just-computed entries \(h'_{\,i+1,c}\) with \(c<c'\), and the two earlier spatial slices \(h'_{\,i,c'}\) and \(h'_{\,i-1,c'}\). Hence a rolling window of three spatial slices suffices, and the update
can be executed block-wise exactly as in
Algorithm~\ref{app:frag_vijp_1d}.

The above derivation generalizes to arbitrary kernel size \(k\). 
In that case, the second summation in \eqref{eq:frag_update_simple} extends over \(j = 1, \dots, k{-}1\), 
and we must retain at least the previous \(k{-}1\) cotangent slices in memory.

Unlike the submersive case, the fragmental checkpointing strategy does not require the spatial stride to exceed the padding (i.e., it does not impose \(s \succ p\)).
However, it does require retaining at least one future cotangent value \(h'\) in memory at each step to enable sequential reconstruction.

\begin{algorithm}[t]
\caption{Fully Parallel VIJP Computation for 2D Convolution}
\label{app:vijp_conv2d}
\begin{algorithmic}[1]
\REQUIRE Gradient tensor $g \in \mathbb{R}^{H' \times W' \times C_{\mathrm{out}}}$,  
         kernel $\mathbf{w} \in \mathbb{R}^{k \times k \times C_{\mathrm{out}} \times C_{\mathrm{in}}}$,  
         stride $s$, padding $p$

\STATE Initialize $h[i, j, c] \gets 0$ for all $(i, j, c)$
\FOR{\textbf{each} output position $(i, j)$ \textbf{in parallel}}
  \FOR{$c = 0$ to $C_{\mathrm{in}} - 1$}
    \STATE $v \gets g[s i, s j, 0] \,/\, \mathbf{w}[p, p, 0, c]$
    \FOR{$c' = 1$ to $C_{\mathrm{out}} - 1$}
      \STATE $v \gets g[s i, s j, c']$
      \FOR{$c'' = 0$ to $c' - 1$}
        \STATE $v \gets v - \mathbf{w}[p, p, c'', c] \cdot h[i, j, c'']$
      \ENDFOR
      \STATE $h[i, j, c'] \gets v$ \hfill \text{$\triangleright$ diagonal is normalized to 1}
    \ENDFOR
  \ENDFOR
\ENDFOR
\RETURN $h$
\end{algorithmic}
\end{algorithm}


\begin{algorithm}
\caption{Fragmental Parallel VIJP Computation for 1D Convolution}
\label{app:frag_vijp_1d}
\begin{algorithmic}[1]
\REQUIRE Gradient tensor $g \in \mathbb{R}^{N \times C_{\mathrm{out}}}$,  
         kernel $\mathbf{w} \in \mathbb{R}^{k \times C_{\mathrm{in}} \times C_{\mathrm{out}}}$,  
         block size $B$,  number of blocks $M$,
         stored cotangents $h_{\mathrm{init}} \in \mathbb{R}^{M(k - 1) \times C_{\mathrm{in}}}$

\FOR{\textbf{each} block $b = 0$ to $M - 1$ \textbf{in parallel}}
  \FOR{$i = k - 1$ to $B - 1$}
    \FOR{$c = 0$ to $C_{\mathrm{in}} - 1$}
      \STATE $v \gets g^{(b)}[i - 1, c]$
      \FOR{$j = 1$ to $k - 1$}
        \STATE $v \gets v - \sum_{c' = 0}^{C_{\mathrm{in}} - 1} \mathbf{w}[j, c', c] \cdot h_{}[i - j, c']$
      \ENDFOR
      \STATE $h_{}[i, c] \gets v$
    \ENDFOR
  \ENDFOR
\ENDFOR
\RETURN $h$
\end{algorithmic}
\end{algorithm}

\section{Complexity Analysis}\label{sec:ans}

\begin{table*}[t]
  \caption{Asymptotic complexity and key characteristics of Moonwalk, its pure variant, and four existing methods, analyzed in~\Cref{sec:ans}. \B{Forward:} only operates in forward-mode; \B{Submersive:}~applicable to non-invertible submersive networks.}
  \centering
  \begin{tabular}{@{}lccccc@{}}
    \toprule
    Method & Time & Memory & High-variance  & Forward  & Submersive\\
    \midrule
    Backprop & $O(n^2L+ndL)$ & $O\left(M_x L + M_\theta L\right) $ & \greencross  & \redcross & \greentick  \ \\
    \makecell[l]{Backprop\\ \, + checkpoint} & $O(n^2L+ndL)$ & $O( \sqrt{n( M_x + M_\theta)L})$ & \greencross  & \redcross & \greentick  \ \\
    Forward-mode & $O(n^2dL^2)$ & $O(M_x + M_\theta)$ & \greencross & \greentick & \greentick \\
    ProjForward & $O(n^2L+ndL)$ & $O(M_x + M_\theta)$ &  \redtick  & \greentick & \greentick  \\ 
    RevBackprop &  $O(n^2L+ndL)$ & $O(M_x + M_\theta)$ & \greencross & \redcross  & \redcross\\ 
    \midrule
    Pure-Moonwalk & $O(n^3L+ndL)$ & $O(M_x + M_\theta)$ & \greencross& \greentick & \greentick\\ 
    Moonwalk &  $O(n^2L+ndL)$ & $O(M_x L + M_\theta)$ & \greencross  & \redcross &\greentick\\
    \makecell[l]{Monwalk\\\, + checkpoint} &  $O(n^2L+ndL)$ & $O( \sqrt{n M_x L} + M_\theta)$ & \greencross & \redcross &\greentick\\

    \bottomrule
  \end{tabular}

  \label{tab:orders}
\end{table*}

While an estimate of the exact time and memory consumption of different methods for computing the gradients can greatly depend on the choice of the network's architecture and the detailed implementation, in this section we will provide an asymptotic analysis, in terms of the main architectural parameters, of the time and memory complexities of our methods and compare them with related previous works~(\Cref{tab:orders}). We omit all methods' linear dependence in time and memory on the mini-batch size. We analyze the computational complexity of the following methods:

\begin{enumerate}
    \item \textbf{Backprop:} Throughout the forward pass (or during a checkpointed block's forward recomputation), all residuals are cached, and subsequently, during a backward pass, gradients for each layer are computed using \textit{vjp}.
    \item \textbf{Forward:} During the forward pass, complete Jacobians for each layer are computed using \textit{jvp}. In practice, a separate forward pass is used for each column, to reuse memory for the same time complexity.
    \item \textbf{ProjForward:} In projected forward~\citep{baydin2022gradients}, parameter gradients projected in a random or predicted direction are obtained using \textit{jvp} concurrently with the forward pass.
    \item \textbf{RevBackprop:} In invertible networks~\citep{revnet}, no residuals need to be stored during the forward pass. In a subsequent backward pass, the output of each layer is used to compute its input via the inverse function, as well as its parameter gradient via \textit{vjp}.
    \item \textbf{Moonwalk:} The input cotangent is computed using Backprop, to reduce computation time at the expense of some memory impact (\Cref{sec:mixed-forward}).
    \item \textbf{Pure-Moonwalk:} Initially, the input cotangent is computed using Forward. Then parameter gradients are obtained using \textit{vijp} and \textit{vjp} in a second forward pass (\Cref{sec:pure-forward}).
\end{enumerate}


We evaluate time based on the standard cost of matrix multiplication, i.e. the product of their two outer dimensions and shared inner dimension, without considering optimization tricks, sparse layers, or other network structures. To evaluate memory, we define $M_{x, i}$ to be the required memory to store the necessary information to compute $\nicefrac{\partial x_i}{\partial x_{i-1}}$, and $M_{\theta, i}$ the added memory to also compute $\nicefrac{\partial x_i}{\partial \theta_i}$. For simplicity, we assume that these values, as well as $n_i$ and $d_i$, scale similarly across layers, and omit the layer index. We refer to memory consumption as the extra amount of memory needed to compute gradients without reflecting the memory to store the parameters or gradients themselves after computation. 

\textbf{Memory complexity}: For Backprop, we have to store residuals required for both input and parameter gradients for every layer, which results in $O(M_{x}L +M_{\theta}L)$ memory complexity. For Moonwalk, we only need to store $M_{x}$ for every layer in the first phase, in order to compute the input cotangent $h_0$, and then we can reuse $M_\theta$ in the second phase after computing each parameter gradient, for a total complexity of $O(M_x L + M_{\theta})$. 
All other methods can discard activations after each layer, for a complexity of $O(M_x + M_\theta)$. 

\textbf{Memory complexity with checkpointing}:
In the case of Backprop with checkpointing, we will have additional memory of $O(cn)$, where $c \le L$ is the number of checkpoints. Then, during backward, we must reconstruct each block of $O(\nicefrac Lc)$ layers and store residuals in $O((M_x+M_\theta)\nicefrac Lc)$ memory. The best trade-off, obtained at $c = O(\sqrt{(M_x+M_\theta)\nicefrac Ln})$ if feasible, is $O( \sqrt{n( M_x + M_\theta)L})$ memory. We can similarly apply checkpointing to the first phase of Moonwalk, which has no need to store $M_\theta$ when reconstructing from a checkpoint, for overall memory of $O( \sqrt{n M_x L} + M_\theta)$. In that case, we still prefer Moonwalk when $M_\theta \gg M_x$, although to a lesser extent than without checkpointing: in the extreme case that layers are so complex that we should checkpoint each one, $nL = O(M_x + M_\theta)$ and both Backprop and Moonwalk require $O(M_x + M_\theta)$ memory. 
On the other hand, any non-negligible $M_\theta$ ensures the benefit of Moonwalk when $L = \omega(\nicefrac{M_\theta}{n})$.


\textbf{Time complexity for Backprop and RevBackprop}: Backprop computation consists of computing two vector–Jacobian products in each layer $i$, $\text{vjp} (f_i, x_{i-1}, h_i)$ and $\text{vjp} (f_i, \theta_i, h_i)$,  which accounts for per-layer time complexity of $O(n^2)$ and $O(nd)$, respectively, and for a total of $O(n^2L+ndL)$ time. RevBackprop additionally needs to evaluate the inverse function $f_i^{-1}(x_i)$, which does not impact the overall complexity in the terms we consider.

\textbf{Time complexity for Forward-mode and ProjForward}:
In Forward-mode, each single parameter $\theta_{j,\ell}$ in layer $j$, of the total $dL$ parameters, generates a pass to compute its gradient, in which we compute $\text{jvp} (f_i, x_{i-1}, \nicefrac{\partial x_{i-1}}{\partial \theta_{j,\ell}})$ with complexity $O(n^2)$ in each layer $i = j+1, \ldots, L$, for a total of $O(n^2 d L^2)$ time.
ProjForward with tangent $u = \{u_i\}_{i=1,\ldots,L}$ is similar to Forward-mode with just a single pass instead of $dL$ passes, but additionally accumulating $\text{jvp} (f_i, \theta_i, u_i)$ in each layer $i$, for a total of $O(n^2 L + ndL)$ time, which coincides with the time complexity of Backprop.

\textbf{Time complexity for Moonwalk and Pure-Moonwalk}: The first phase of Pure-Moonwalk computes $\text{jvp} (f_i, x_{i-1}, \frac{\partial x_{i-1}}{\partial x_0}e_\ell)$ in each layer $i$ for each input element $\ell$, for a total time complexity of $O(n^3L)$. The second phase computes $\text{vijp} (f_i, x_{i-1}, h_{i-1})$ and $\text{vjp} (f_i, \theta_i, h_i)$ in each layer for $O(n^2 L + ndL)$ more, and a total of $O(n^3 L + ndL)$ time.
Moonwalk replaces the first phase with Backprop for just the input gradient, incurring time complexity $O(n^2L)$. Together with the same second phase as in Moonwalk, this totals $O(n^2 L + ndL)$ time complexity.

\end{document}